\documentclass{article} 
\usepackage{iclr2019_conference,times}

\usepackage{amsmath,amsfonts,bm}

\def\eqref#1{equation~\ref{#1}}

\def\1{\bm{1}}

\def\ra{{\textnormal{a}}}

\def\rx{{\textnormal{x}}}

\def\rva{{\mathbf{a}}}

\def\erva{{\textnormal{a}}}

\def\ervx{{\textnormal{x}}}

\def\rmA{{\mathbf{A}}}

\def\vmu{{\bm{\mu}}}
\def\vtheta{{\bm{\theta}}}
\def\va{{\bm{a}}}

\def\ve{{\bm{e}}}

\def\vx{{\bm{x}}}

\def\eva{{a}}

\def\mA{{\bm{A}}}

\def\mH{{\bm{H}}}
\def\mI{{\bm{I}}}
\def\mJ{{\bm{J}}}

\def\mX{{\bm{X}}}

\def\mSigma{{\bm{\Sigma}}}

\DeclareMathAlphabet{\mathsfit}{\encodingdefault}{\sfdefault}{m}{sl}
\SetMathAlphabet{\mathsfit}{bold}{\encodingdefault}{\sfdefault}{bx}{n}
\newcommand{\tens}[1]{\bm{\mathsfit{#1}}}
\def\tA{{\tens{A}}}

\def\tX{{\tens{X}}}

\def\gG{{\mathcal{G}}}

\def\sA{{\mathbb{A}}}
\def\sB{{\mathbb{B}}}

\def\sS{{\mathbb{S}}}

\def\emA{{A}}

\newcommand{\etens}[1]{\mathsfit{#1}}

\def\etA{{\etens{A}}}

\newcommand{\E}{\mathbb{E}}

\newcommand{\R}{\mathbb{R}}

\newcommand{\KL}{D_{\mathrm{KL}}}
\newcommand{\Var}{\mathrm{Var}}

\newcommand{\Cov}{\mathrm{Cov}}

\newcommand{\normltwo}{L^2}
\newcommand{\normlp}{L^p}

\newcommand{\parents}{Pa} 

\usepackage{hyperref}
\usepackage{url}
\usepackage{graphicx}

\title{Generalized Capsule Networks with \\Trainable Routing Procedure}

\author{Zhenhua Chen  \\
School of School of Informatics, \\Computing, and Engineering\\
Indiana University\\
Bloomington, IN 47408, USA \\
\texttt{\{chen478\}@iu.edu} \\
\And
David Crandall  \\
School of School of Informatics, \\Computing, and Engineering\\
Indiana University\\
Bloomington, IN 47408, USA \\
\texttt{\{djcran\}@iu.edu} \\
}

\iclrfinalcopy 
\begin{document}

\maketitle

\begin{abstract}
CapsNet (Capsule Network) was first proposed by~\citet{capsule} and later another version of CapsNet was proposed by~\citet{emrouting}. CapsNet has been proved effective in modeling spatial features with much fewer parameters. However, the routing procedures in both papers are not well incorporated into the whole training process. The optimal number of routing procedure is misery which has to be found manually.  To overcome this disadvantages of current routing procedures in CapsNet, we embed the routing procedure into the optimization procedure with all other parameters in neural networks, namely, make coupling coefficients in the routing procedure become completely trainable. We call it Generalized CapsNet (G-CapsNet). We implement both ``full-connected" version of G-CapsNet and ``convolutional" version of G-CapsNet. G-CapsNet achieves a similar performance in the dataset MNIST as in the original papers. We also test two capsule packing method (cross feature maps or with feature maps) from previous convolutional layers and see no evident difference. Besides, we also explored possibility of stacking multiple capsule layers. 
The code is shared on \hyperlink{https://github.com/chenzhenhua986/CAFFE-CapsNet}{CAFFE-CapsNet}.
\end{abstract}

\section{Introduction}
According to ~\citet{capsule}, a capsule is a group of neurons whose activity vector represents the instantiation parameters of a specific type of entity such as an object or an object part. Intuitively, CapsNet can better model spatial relationship by using much more fewer parameters. The experiments in~\citet{capsule} proved this argument in some small datasets like MNIST~\citet{} and smallNORB~\citet{}. However, the routing procedure is very computationally expensive, and the number of routing iterations has to be manually set by testing. To overcome this issue, we propose Generalized CapseNet. The key idea of CapsNet is incorporating the routing procedure to the overall optimization procedure. In other words, it makes the coupling coefficients trainable instead of being calculated by dynamic routing~\citet{capsule} or EM routing~\citet{emrouting}. 

Another interesting question yet to answer is that how to package a capsule from activations of previous convolutional layers. We can select the elements if each capsule at the same position across different feature maps (as Figure~\ref{package} Left shows). We can also select the elements of each capsule within each feature map (for example, we can choose each row or column as a capsule as Figure~\ref{package} Right shows). Two methods seem different, and the previous one makes more sense since capsules are supposed to capture different spatial features. However, according to our experiment, both ways have similar performance. We do not know the exact reason for the question but will try to answer it in the experiment session.

CapsNet seems more general than standard neural networks (neurons can be considered as a unit length of capsules), so we also explore the scalability of CapsNet by building a two-layer G-CapsNet. Unfortunately, the network is inclined to get saturated even after a capsule version of ReLU layer is added. The same thing happens when we try to `capsulize' the whole network, namely capsules other than neurons are the atomic units. For example, a colored input image can be considered as $h\times w$, 3-dimension capsules, so are the following layers.

\section{Generalized CapsNet}
To better illustrate the idea of G-Capsnet, let's check what the normal neural networks' loss function looks like. Assume we have a dataset $\{ (x^{(1)}, y^{(1)}), \ldots, (x^{(m)}, y^{(m)}) \}$. If we can use neural networks to train a model, the loss function can be defined as Equation~\ref{eq:1}: 

\begin{equation} \label{eq:1}
    J(W,b) = \left[ \frac{1}{m} \sum_{i=1}^m \left( \frac{1}{2} \left\| h_{W,b}(x^{(i)}) - y^{(i)} \right\|^2 \right) \right]
                       + \frac{\lambda}{2} \sum_{l=1}^{n_l-1} \; \sum_{i=1}^{s_l} \; \sum_{j=1}^{s_{l+1}} \left( W^{(l)}_{ji} \right)^2
\end{equation}
The loss function for G-CapsNet is similar. The only difference is we have extra coupling coefficients (through routing procedure) to train. Note that $\left\| h_{W,b}(x^{(i)}) \right\|$ is the mathematical form of the neural network that we assume would fit the dataset. $n_l$, $s_l$, $s_{l+1}$ are the number of layers, the number of neurons of layer $l$ and the number of neurons of layer $l+1$. $W$ and $b$ are the parameters we are supposed to get by training the neural network. 

For the generalized CapsNet, $W^{(l)}_{ji}$ is the transformation matrix that maps one type of capsule in one layer to another type of capsule on top of it. $ c^{(l)}_{ji}$ is the coupling coefficient between adjacent layers. 
\begin{equation} \label{eq:2}
    J(W,b,c) = \left[ \frac{1}{m} \sum_{i=1}^m \left( \frac{1}{2} \left\| h_{W,b,c}(x^{(i)}) - y^{(i)} \right\|^2 \right) \right]
                       + \frac{\lambda}{2} \sum_{l=1}^{n_l-1} \; \left(\sum_{i=1}^{s_l} \; \sum_{j=1}^{s_{l+1}} \left( W^{(l)}_{ji} \right)^2 + 
                       \sum_{i=1}^{p_l} \; \sum_{j=1}^{p_{l+1}} \left( c^{(l)}_{ji} \right)^2 \right)
\end{equation}

\subsection{Structure of Generalized CapsNet}
\begin{figure}
  \centering
  \includegraphics[width=0.6\linewidth]{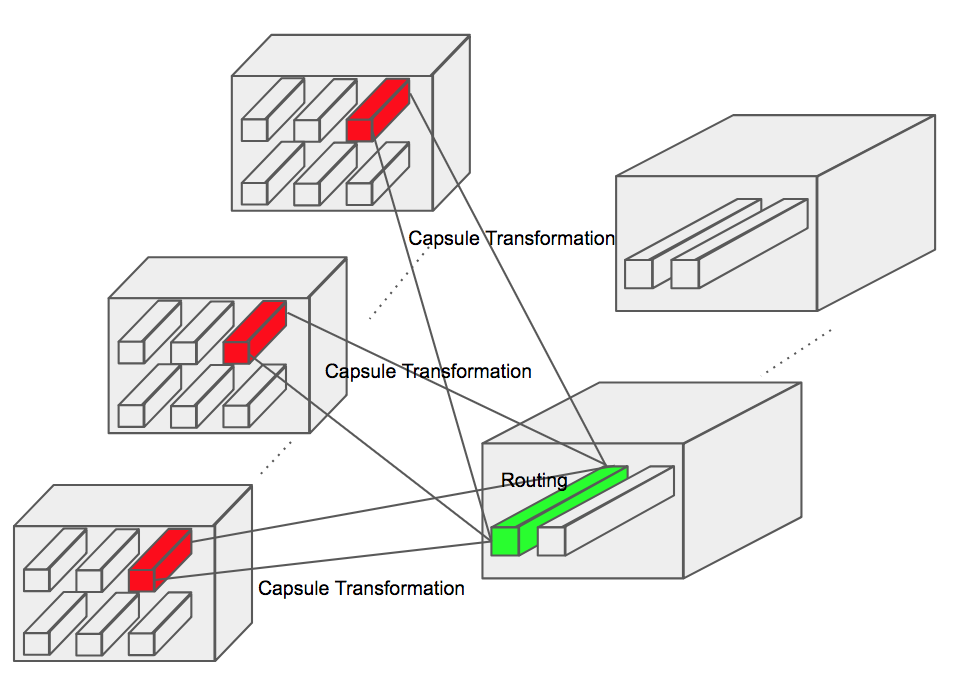}
  \caption{The structure of generalized CapsNet.}
  \label{capsnet}
\end{figure}

Like the structure of the previous CapsNets, the generalized CapsNet also includes two phases, capsule transformation, and capsule routing, as Figure~\ref{capsnet} shows. Capsule transformation converse one type of capsules into another type of capsules. For example, \citet{capsule} transforms 8-dimension capsules into 16-dimension capsules while \citet{emrouting} transforms $4\times4$capsules into $4\times4$ capsules. In theory, we can transform type of capsules into any other type of capsules. The capsules can be of any shape (vector, matrix, cube or even hyper-cube). Assume the shape of capsules $\mathbf{u}$ in the lower layer is $(m_1, m_2, \dots, m_k)$, the transformation matrix $\mathbf{W}$ is $m_k, n_1, n_2, \dots, n_k$, then the shape of output capsules $\mathbf{v}$ in the higher layer is $m_1, m_2, \dots, m_k, n_1, n_2, \dots, n_k$. Capsule routing ensures capsules in lower layers are scaled and sent to their parent capsules in higher layers, as Equation~\ref{eq:3} shows.

\begin{equation} \label{eq:3}
    \mathbf{v_j} = \sum_{i}c_{ij}\mathbf{u_{j|i}}, \quad\quad \mathbf{u_{j|i}} = \mathbf{W_{ij}}\mathbf{u_{i}}
\end{equation}

\subsection{Squash Function}
We adopt two versions of squash functions which are proposed by~\citet{capsule} and ~\citet{squash2}. As Equation~\ref{eq:4} shows. Squash functions here equal to a normalization operation which also bring non linearity. 
\begin{equation} \label{eq:4}
    \mathbf{v_j^{'}} = \frac{\mathbf{\|v_j\|}^2}{1 + \mathbf{\|v_j\|}^2}\frac{\mathbf{v_j}}{\|\mathbf{v_j}\|} \quad\quad \mathbf{v_j^{'}} = \left(1 - \frac{1}{\mathbf{e}^{\|\mathbf{v_j}\|}}\right) \frac{\mathbf{v_j}}{\|\mathbf{v_j}\|}
\end{equation}

\subsection{Loss Function}
We adopt the same margin loss function in~\cite{capsule}, as Equation~\ref{eq:5} shows.

\begin{equation} \label{eq:5}
    L_k=T_k\max(0, m^+-\|\mathbf{v_k}\|)^2 + \lambda (1-T_k) \max(0, \|\mathbf{v_k}\| - m^-)^2
\end{equation}

\section{Experiments}
\subsection{full connected G-CapsNet on MNIST}
We adopt the same baseline as in paper~\citet{}. The first convolutional layer outputs 256 feature maps. The second convolutional layer outputs 256 feature maps or 32$\times$6$\times$6 8\-D capsules. The last two layers are fully connected layers. Please check our released code or the original paper written by~\citet{capsule} for more details. 

We call the Capsule structure that~\citet{capsule} proposed full connected CapsNet since each capsule in the higher layer connects to every capsule in the lower layer. As Table~\ref{tab:1} shows, no matter whether the reconstruction procedure involved, G-CapsNet can always achieve better performance by using much less number of parameters. Note that the performance of the baseline and the G-CapsNet reported here is a little lower than in~\citet{capsule}, we consider the difference is caused by different deep learning frame (We use Caffe other than TensorFlow). 

\begin{figure}
  \centering
  \includegraphics[width=0.4\linewidth]{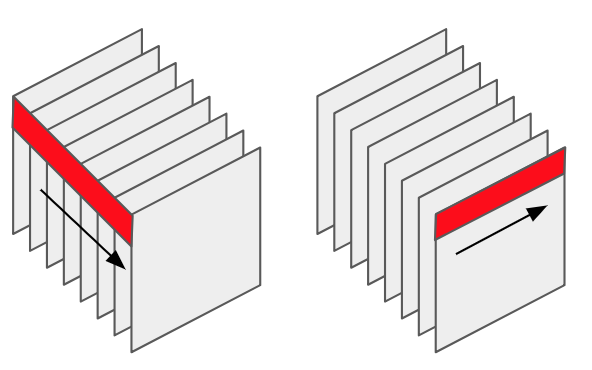}
  \caption{Packaging capsules across feature maps VS within feature maps}
  \label{package}
\end{figure}

Another interesting thing we found is that the way of stacking the activation of neurons does not matter a lot (0.68 VS 0.66). It makes more sense to package capsules across different feature maps since the capsules are supposed to capture a couple of different types of features at the same position, as Figure~\ref{package} left shows. However, what we found is that packaging capsules within each feature map (as Figure~\ref{package} right shows) as opposed to across feature maps can achieve similar performance. The potential reason for this is that no matter how we package capsules, once the organization of these capsules is fixed, the network will finally learn potential spatial relationship.

\subsection{Convolutional G-CapsNet on MNIST}
Similarly, we call the structure in ~\citet{emrouting} as convolutional CapsNet since the same type of capsules of different positions shares the same transformation matrices. The convolutional G-CapsNet we use here is similar to the previously mentioned full connected G-CapsNet except we use 6 by 6 kernel for calculating the  and 4 by 4 ``matrix" capsule instead of 1 by 16 ``vector" capsule for the capsule layer. Please check our project for more details. As Table~\ref{tab:1} shows, convolutional G-CapsNet can achieve better performance compared to the baseline by using fewer parameters. \\

Above G-CapsNets all have a single capsule layer, then a natural question arises, can we build a Multi-layer G-CapsNet? After all, neural networks can be considered as a special type of CapsNet (the length of each capsule is 1). One straightforward way is stacking multiple capsule layers on top of each other. However, we found this type of CapsNet is easy to get saturated. To solve this issue, we design a capsule version of ``ReLU" layer which makes the performance better but still far from satisfying. How to make a CapsNet more scalable will be our next work. 

\begin {table}
\begin{center} \label{tab:1} 
\begin{tabular}{ |c|c|c|c| } 
\hline
Algorithm & error rate(\%) & number of parameters \\
baseline & 0.83 & 35.4M\\ 
full connected G-CapsNet & 0.66 & 8.2M\\ 
full connected G-CapsNet* & 0.66  &6.8M\\
full connected G-CapsNet**  & 0.76 & 8.2M\\
Convolutional G-CapsNet &  0.75 & 6.9M\\ 
Convolutional G-CapsNet* & 0.70 & 5.5M\\ 

\hline
\end{tabular}
\caption{Error rate VS number of parameters on MNIST. Note that ``*" means no reconstruction, ``**" means the alternate way of stacking capsules.}
\end{center}
\end{table}

\section{Conclusion}
G-CapsNet incorporates the routing procedure of capsules into the whole optimization process which gets rid of the set of routing times and guarantees the convergence. Also, we tried two different ways of packaging capsules and conclude that how to packaging capsules does not matter a lot. Finally, we evaluated muli-layer CapsNets and found that multi-layer CapsNets are easy to get saturated. How to make CapsNet scalable is still an open question.

\bibliography{iclr2019_conference}
\bibliographystyle{iclr2019_conference}

\end{document}